\documentclass[a4paper,twocolumn,10pt]{article}
\usepackage{hyperref}       
\usepackage{usenix}

\usepackage[utf8]{inputenc} 
\usepackage[T1]{fontenc}    
\usepackage{url}            
\usepackage{booktabs}       
\usepackage{amsfonts}       
\usepackage{nicefrac}       
\usepackage{microtype}      
\usepackage{lipsum}
\usepackage{titlesec}
\usepackage{graphicx}
\usepackage[colorinlistoftodos]{todonotes}
\usepackage{breakurl}
\usepackage{setspace}
\usepackage{float}
\usepackage{wrapfig}
\usepackage[font={footnotesize, it}, skip=-12pt]{caption}
\usepackage{chngpage}

\usepackage{appendix}
\usepackage{svg}

\newcounter{qcounter}
\title{Understanding (Non-)Robust Feature Disentanglement and the Relationship Between Low- and High-Dimensional Adversarial Attacks}

\author{
  \rm{Zuowen Wang, Leo Horne}\\
  \normalfont\texttt{\{wangzu, hornel\}@ethz.ch} \\
}

\begin{document}
\setstretch{1.0}
\maketitle

\newcommand{\makepurple}[1]{{\color{purple}{#1}}}
\newcommand{\makered}[1]{{\color{red}{#1}}}
\newcommand{\makeblue}[1]{{\color{blue}{#1}}}
\newcommand{\makegreen}[1]{{\color{green}{#1}}}
\newcommand{\nmcomment}[1]{\makered{{NM: $\quad$ #1}}}

\newcommand{\newres}[1]{{\color{green}{#1}}}
\newcommand{\newzw}[1]{{\color{red}{#1}}}

\newcommand{\fycomment}[1]{\makered{{FY: $\:\:$ #1}}}
\newcommand{\chcomment}[1]{\makepurple{{CH: #1}}}
\newcommand{\zwcomment}[1]{\makegreen{{ZW: #1}}}

\newcommand\Christina[1]{\chcomment{#1}}
\newcommand{\searchset}[1]{S^{#1}}
\newcommand{\Trans}[1]{L_{#1}}
\newcommand{\param}{\theta}
\newcommand{\px}{\textrm{px}}
\newcommand{\core}{\ensuremath{\text{CoRe}}}
\newcommand{\EE}{\mathbb{E}\:}
\newcommand{\ID}{\ensuremath{\text{ID}}}
\newcommand{\attvec}{\ensuremath{Z}}
\newcommand{\corepen}{\ensuremath{\phi}}
\newcommand{\Pert}{\ensuremath{\Delta}}
\newcommand{\Pertsetorig}{\widetilde{\mathcal{S}}}
\newcommand{\Pertset}{\mathcal{S}}
\newcommand{\Trafo}[2]{\mathcal{T}(#1, #2)}
\newcommand{\XID}[1]{\ensuremath{X^{#1}}}
\newcommand{\YID}{\ensuremath{Y}}
\renewcommand{\P}{\mathbb{P}}
\newcommand{\loss}{\ell}
\newcommand{\func}{f}
\newcommand{\cond}{\: | \:}
\newcommand{\Var}{\text{Var}}
\newcommand{\numid}{n}
\newcommand{\reg}{R}
\newcommand{\Loss}[1]{\mathcal{L}_{#1}}
\newcommand{\PopLoss}[1]{\Loss{#1}}
\newcommand{\PopLossReg}[1]{\PopLoss{#1}}
\newcommand{\EmpLoss}[1]{\widehat{\mathcal{L}}_{#1}}
\newcommand{\stepsize}{\alpha}
\newcommand{\adv}{\text{adv}}
\newcommand{\mixadv}{\text{mix}}
\newcommand{\Xadv}[2]{X^{\star}_{#2}(\param^{#1})}
\newcommand{\Xadvce}[2]{X^{\text{ce}}_{#2}(\param^{#1})}
\newcommand{\Xadvreg}[2]{X^{\Reg}_{#2}(\param^{#1})}

\newcommand{\RN}{\mathbb{R}}
\newcommand{\batchsize}{b}
\newcommand{\batch}{B}
\newcommand{\gradient}{G}
\newcommand{\groupsize}{\numadv}
\newcommand{\X}{\mathcal{X}}
\newcommand{\paramopt}{\param^{\star}}
\newcommand{\Indi}[1]{\mathbb{I}\{#1\}}
\newcommand{\prederr}{R}
\newcommand{\predacc}{A}
\newcommand{\dimparam}{p}
\newcommand{\qid}{q}
\newcommand{\epsadv}{Q}
\newcommand{\probpert}{P}
\newcommand{\acronym}{RECOVARIT }
\newcommand{\acronymmath}{\text{RECOVARIT}}
\newcommand{\ibatch}{I}
\newcommand{\ul}[1]{\underline{#1}}
\newcommand{\mutual}[2]{I(#1;#2)}
\newcommand{\dimid}{m}
\newcommand{\attstyle}{Z^{\text{style}}}
\newcommand{\attcore}{Z^{\text{core}}}
\newcommand{\Pertadv}{\Pert^{\star}}
\newcommand{\dimpert}{q}
\newcommand{\numadv}{M}
\newcommand{\EmpVar}{\widehat{\Var}}
\newcommand{\EmpEE}{\widehat{\mathbb{E}}}
\newcommand{\EEt}{\mathbb{E}^t}
\newcommand{\updateop}{M}
\newcommand{\full}{\text{100}}
\newcommand{\half}{\text{50}}
\newcommand{\mixb}{\half}
\newcommand{\natb}{\text{0}}
\newcommand{\advb}{\full}
\newcommand{\quarter}{\text{(25)}}

\newcommand{\std}{\ensuremath{\text{std}}}
\newcommand{\stdaug}{\ensuremath{\std^\star}}
\newcommand{\rnd}{\text{rnd}}
\newcommand{\recovarit}[3]{\text{#1} (#3) \text{ w/ } #2}
\newcommand{\at}{\text{Adversarial training}}
\newcommand{\atabb}{\text{Adv.\ Training}}
\newcommand{\wokabb}{\text{Wo-$\numwok$}}
\newcommand{\woabb}[1]{\text{Wo-${#1}$}}
\newcommand{\wok}{\text{worst-of-$\numwok$}}
\newcommand{\wo}[1]{\text{worst-of-${#1}$}}
\newcommand{\wokfull}{\text{Worst-of-$\numwok$ \full}}
\newcommand{\wofull}[1]{\text{Worst-of-${#1}$ \full}}
\newcommand{\wokhalf}{\text{Worst-of-$\numwok$ \half}}
\newcommand{\wohalf}[1]{\text{Worst-of-${#1}$ \half}}
\newcommand{\numwok}{k}
\newcommand{\fo}{\text{S-PGD}}
\newcommand{\fofull}{\fo \text{ \full}}
\newcommand{\fohalf}{\fo \text{ \half}}
\newcommand{\fstar}{f^\star}
\newcommand{\suppmat}{Appendix}
\newcommand{\defense}{$D$}
\newcommand{\defvar}{\texttt{def}}
\newcommand{\dataug}{DA}
\newcommand{\eps}{\epsilon}
\newcommand{\paramhat}{\widehat{\param}}
\newcommand{\numtotal}{N}
\newcommand{\lossbound}{B}
\newcommand{\invfuncs}{\mathcal{F}_{\text{inv}}}
\newcommand{\empinvfuncs}{\mathcal{F}_{\numid, \text{inv}}}
\newcommand{\dudley}{D(\func, \lossbound)}
\newcommand{\cover}{N}
\newcommand{\funcloss}{\mathcal{F}_\ell}
\newcommand{\funcspace}{\mathcal{F}}
\newcommand{\paramhatopt}{\paramhat^{\star}}
\newcommand{\atmix}{\ensuremath{\atabb\:(50)}}
\newcommand{\atfull}{\ensuremath{\atabb\:(100)}}
\newcommand{\recovaritmix}{\ensuremath{\acronymmath\:(50)}}
\newcommand{\recovaritfull}{\ensuremath{\acronymmath\:(100)}}
\newcommand{\Rade}{\mathcal{R}}
\newcommand{\E}{\text{e}}
\newcommand{\paramspace}{\Theta}
\newcommand{\Group}{\mathbb{G}}
\newcommand{\defn}{:=}
\newcommand{\Gorbit}[1]{\Group(#1)}
\newcommand{\SubGorbit}[1]{G^{#1}}
\newcommand{\SubGroup}[1]{\Group^{#1}}
\newcommand{\ZOLoss}[1]{\mathcal{R}_{#1}}
\newcommand{\xworst}[1]{x^\star({#1})}
\newcommand{\allorbits}{\mathcal{G}}
\newcommand{\ortho}{\perp \!\!\! \perp }
\newcommand{\InvF}{\mathcal{V}}
\newcommand{\Fspace}{\mathcal{F}}
\newcommand{\fbest}{f^\star}
\newcommand{\lip}{\gamma}
\newcommand{\nat}{\text{nat}}
\newcommand{\mix}{\text{mix}}
\newcommand{\fhat}{\hat{f}}
\newcommand{\Reg}{\ensuremath{R}}
\newcommand{\fhatt}{\hat{f}'}
\newcommand{\RegFun}{R_{\monfunc}}
\newcommand{\monfunc}{h}
\newcommand{\Pertworst}{\Pert^\star}
\newcommand{\frob}{f^{\text{rob}}}
\newcommand{\abest}{a^\star}
\newcommand{\Pertdiam}{\Pert^{\star}}
\newcommand{\regfunc}{h}
\newcommand{\regadvnat}{\tilde{R}}
\newcommand{\AT}{\text{AT}}
\newcommand{\ATM}{\text{ATM}}
\newcommand{\ALP}{\text{ALP}}
\newcommand{\tr}{\text{KL}}
\newcommand{\trcce}{\text{KL-C}}
\newcommand{\ltwo}{\ensuremath{\ell_2}}
\newcommand{\linf}{\ensuremath{\ell_{\infty}}}
\newcommand{\alp}{ALP}
\newcommand{\trades}{\text{TRADES}}
\newcommand{\orbitset}{\mathbb{G}}
\newcommand{\KL}{\text{D}_{\text{KL}}}
\newcommand{\advtype}{\texttt{adv}}
\newcommand{\stn}{\text{STN}}
\newcommand{\grn}{\text{GRN}}
\newcommand{\fnat}{\ensuremath{f^{\text{nat}}}}
\newcommand{\advex}{\texttt{examples}}
\newcommand{\batchtype}{\texttt{batch}}
\newcommand{\advreg}{\texttt{\Reg}}
\newcommand{\advcce}{\text{ce}}
\newcommand{\advboth}{\ensuremath{B}}
\newcommand{\regtype}{\texttt{Reg}\:}
\newcommand{\cce}{\text{CE}}
\newcommand{\condind}{\purp}
\newcommand{\gresnet}{\text{G-ResNet44}}
\newcommand{\etn}{\text{ETN}}
\newcommand{\aug}[1]{\ensuremath{{#1}^{\star}}}
\newcommand{\twostage}{\text{STN+}}
\newcommand{\ZN}{\mathbb{Z}}
\newcommand{\Gspace}{\mathcal{I}}
\newcommand{\SetOrbit}{\mathcal{G}}
\newcommand{\orbitfunc}{\Psi}
\newcommand{\SubSetOrbit}{\widetilde{\SetOrbit}}
\newcommand{\Frob}{F^{\rob}}
\newcommand{\rob}{\text{rob}}
\newcommand{\fmin}{f^{\min}}
\newcommand{\reglosstype}{\texttt{r-loss}}
\newcommand{\natreg}{\natb}
\newcommand{\robreg}{\full}
\newcommand{\robregmix}{\half}
\newcommand{\Xsupp}{\widetilde{\X}}

\newcommand\blfootnote[1]{%
  \begingroup
  \renewcommand\thefootnote{}\footnote{#1}%
  \addtocounter{footnote}{-1}%
  \endgroup
}

\begin{abstract}

Recent work has put forth the hypothesis that adversarial vulnerabilities in neural networks are due to them overusing ``non-robust features'' inherent in the training data. We show empirically that for PGD-attacks, there is a training stage where neural networks start heavily relying on non-robust features to boost natural accuracy. We also propose a mechanism reducing vulnerability to PGD-style attacks consisting of mixing in a certain amount of images containing mostly ``robust features'' into each training batch, and then show that robust accuracy is improved, while natural accuracy is not substantially hurt. We show that training on ``robust features'' provides boosts in robust accuracy across various architectures and for different attacks. Finally, we demonstrate empirically that these ``robust features'' do not induce spatial invariance.

\end{abstract}



\section{Introduction}\label{sec:intro}

In recent years, deep neural networks (DNNs) have become the tool of choice for image classification tasks. State-of-the-art DNN models are able to achieve very high accuracies on standard datasets. However, the complexity of these neural networks has resulted in human beings being unable to fully understand their training and inference process, causing difficulties in interpreting certain counterintuitive phenomena. One such phenomenon is the existence of \textit{adversarial examples} \cite{Biggio2013, Szegedy13}, images that appear to humans to be obvious examples of a certain class $y_{true}$ of objects (e.g. $y_{true}$~=~`cat'), but that the model assigns to a completely different class $y_{false}$ with very high confidence (e.g. $y_{false}$~=~`dog' with 99\% confidence). In some cases, even simple low-dimensional transformations such as translation or rotation can cause misclassification \cite{Engstrom17}. The existence of adversarial examples \cite{Goodfellow14,Madry18,Engstrom17}, as well as numerous defense schemes \cite{Zhang19, Wong18, Ragunathan18, Engstrom18, Kannan18}, has been well-studied. However, the mechanism underlying their existence is presently not known. 

Recent work by Ilyas et al. \cite{Ilyas2019} has put forth the hypothesis that the emergence of adversarial examples is due to the supervised learning paradigm currently employed, which solely aims to maximize categorical accuracy on natural (unperturbed) images. More specifically, they propose that each image class $y$ may have two main kinds of features associated with it: \textit{robust} features, features that are clearly indicative of class $y_{true}$ both to humans and to neural networks (e.g. the existence of fur or pointed ears), and \textit{non-robust} features, features that are strongly indicative of the class $y_{true}$, but are only meaningful to the neural network and not to humans, to whom they may appear to be random patterns. While non-robust features aid in boosting classification accuracy, they are \textit{brittle}, meaning that slight perturbations to the image (imperceptible to humans) can completely transform the non-robust features to indicate a different class $y_{false}$. The core idea is that such perturbations to non-robust features are the root cause of the existence of adversarial examples, and the dominant supervised learning paradigm causes neural networks to make heavy use of non-robust features to maximize natural accuracy. Ilyas et al. claim that this is the reason adversarial attacks tend to transfer well among diverse neural network architectures.

Ilyas et al. constructed two special datasets to verify their hypothesis: from a base dataset $D$ containing natural images, they created a dataset $D_R$ containing images whose non-robust features cannot be relied on to indicate their class (i.e., their only useful features are robust) and a dataset $D_{NR}$ containing images whose only useful features are non-robust. $D_R$ is produced by using an adversarially-trained classifier $f_\mathcal{A}$ to distort a randomly selected image $x'$ to the label $y$ of a different image $x$. The intuition is that since $f_\mathcal{A}$ is adversarially trained, it will only distort the robust features of $x'$ to the label $y$, leaving the non-robust features untouched and thus not relevant to the class $y$. $D_{NR}$ is produced by constructing adversarial examples (whose non-robust features, by definition, point to a class $y_{false} \neq y_{true}$, but whose robust features point to $y_{true}$) and relabeling them to class $y_{false}$.

More research is required to fully understand if Ilyas et al.'s hypothesis holds, and if so, how and to what it applies. This project aims to clarify and extend some aspects of Ilyas et al.'s work by asking the following questions:

\refstepcounter{qcounter}
\subsection*{Question \arabic{qcounter}.\quad \rm{\textit{Does the current supervised training paradigm induce undesired utilization of non-robust features?}}} \label{q1}

We first ask whether it is possible that the current goal of maximizing a classifier's accuracy during training can lead to the classifier ``overfitting'' to non-robust features, i.e. primarily making use of robust features in the early stages of training but heavily relying on non-robust features later to bring classification accuracy to its peak level. We design an experiment that snapshots training progress at regular intervals, performs various adversarial attacks, and then calculates the attack success rate ($ASR~=~\frac{\rm{number~of~adversarial~examples~wrongly~classified}}{\rm{number~of~corresponding~natural~examples~correctly~classified}}$). The intuition is that if neural networks first learn mainly robust features and only overfit to non-robust features in later training stages, the ASR should be lower in earlier stages of training.

\refstepcounter{qcounter}
\subsection*{Question \arabic{qcounter}.\quad \rm{\textit{Can we further validate the idea of natural images containing both robust and non-robust features?}}} \label{q2}

To answer this question, we design a dataset $D_{mix}$ containing (1) natural images of various classes from the original dataset $D$ and (2) images from $D_R$. We train various classifiers on this dataset and evaluate their robust accuracies for various different attacks as well as their natural accuracies. The intuition is that higher proportions of images from $D_R$ will lead to higher robust accuracies. Also, to eliminate the concern of defense information leakage to the constructed robust dataset \cite{engstrom2019a}, we evaluate the models trained on $D_{mix}$ with different attack methods. We expect that the robustness induced by $D_{mix}$ transfers well across different attack methods and architectures.

\refstepcounter{qcounter}
\subsection*{Question \arabic{qcounter}.\quad \rm{\textit{Can PGD-robust datasets induce spatial equivariance?}}} \label{q3}

Ilyas et al. only examine projected gradient descent (PGD) attacks \cite{Madry18} as a means of adversarial attack. We will therefore examine whether the ``robust features'' of \cite{Ilyas2019} are also robust against spatial attacks (such as translation and rotation). In particular, if such features exist, they should (by definition) induce a certain degree of equivariance to rotation and translation attacks for the model.

To test this hypothesis, we train classifiers on $D_R$ and perform spatial transformation attacks \cite{Engstrom17}. Since the adversarially-trained classifier $f_\mathcal{A}$ used to construct $D_R$ is only known to be robust to PGD attacks \cite{Madry18}, we also attempt to construct spatially robust and non-robust datasets $D_R^S$ and $D_{NR}^S$ using the same method as for $D_R$ and $D_{NR}$, but using special equivariance-inducing classifiers as $f_\mathcal{A}$. 

Finally, we train networks which claim to achieve a certain degree of spatial equivariance on $D_R$ to evaluate whether training only on robust features causes these networks to achieve a higher degree of spatial equivariance compared to training on natural images. The architectures we use include the group equivariant network \cite{Cohen16}, spatial transformer network \cite{Jaderberg15}, and a special case of equivariance transformer models \cite{Esteves18}, the polar transformer network \cite{Tai19}.

\textbf{Contributions.} Concretely, our contributions in this work are the following insights into the claims made in \cite{Ilyas2019}: 
\vspace{-7pt}
\begin{itemize}
    \itemsep-0.3em
    \item We show that the current training goal of maximizing accuracy on natural images is not in itself responsible for overfitting to non-robust features. In fact, we show that neural networks do not make a clear distinction between robust and non-robust features during training, utilizing both from the very beginning of training.
    \item Previous work \cite{Zhang19} has shown that there is often a trade-off between natural accuracy and robust accuracy. We show empirically that we can achieve a boost in robust accuracy by adding only a small amount of images from $D_R$ to the training dataset without perceptibly hurting natural accuracy. Furthermore, we show that the robustness of classifiers trained on $D_R$ transfers well among attacks based on different high-dimensional perturbations as well as among different architectures. We can therefore consider such mixed datasets as a novel defense mechanism against adversarial attacks.
    \item We attempt to apply the theories from \cite{Ilyas2019} to the case of spatial attacks, and find that their theories do not apply in the realm of low dimensional attacks. In particular, classifiers trained on $D_R$ exhibit worse spatial robustness than classifiers trained on $D$, even in the case of spatially robust architectures.
\end{itemize}

\section{Models and Methods}

In this section, we go over the methods used to answer the Questions \ref{q1}, \ref{q2}, and \ref{q3}.

\subsection{Attacks} \label{sec:attacks}
In our work, we mainly study two major classes of attacks (i.e. image perturbations): PGD attacks and spatial 
attacks. The attack settings largely mirror those in \cite{Ilyas2019, engstrom2019a}. We mainly use these attacks for evaluation. The accuracy of a model evaluated with adversarial examples generated by an attack scheme $\mathcal{A}$ is called \textit{robust accuracy} under $\mathcal{A}$, while the original evaluation using only natural images is termed \textit{natural accuracy}. We use various attack parameters for the specific purpose of evaluating degrees of robustness for the different models trained. The character of a model which enables it to resist PGD or spatial attacks is called \textit{PGD robustness} or \textit{spatial robustness} respectively.

\paragraph{PGD attacks} Projected Gradient Descent (PGD) attacks are considered to be the strongest attacks which utilize the first-order information of the network, as studied in \cite{Madry18}. We use two types of PGD attacks (\ltwo-PGD and \linf-PGD) which adopt different norms when bounding the attack space: \ltwo-norm and \linf-norm. The adversarial examples are generated by perturbing the original images towards the gradient ascent direction, while the gradient is normalized by the constraint norm. The resulting perturbed image will be projected back to the allowed image space, with the norm of its difference from the original image constrained by a parameter $\epsilon$. This attack step can be performed for multiple iterations.

For the \ltwo-PGD attack, our attack parameters follow \cite{Ilyas2019}. We choose $\epsilon=0.25$ or $0.5$, step size $=0.1$ and we run 100 iterations to ensure that it is possible for the attack step to reach the boundary. For the \linf-PGD attack, apart from following the attack settings in \cite{Madry18}, we also set up experiments of smaller \linf boundary, i.e. weaker attacks, for our own purposes of demonstration. In total, we have $\epsilon = \frac{1}{255}, \frac{2}{255}, \frac{4}{255}$, and $\frac{8}{255}$ with step size $=\frac{\epsilon}{4}$ and 7 steps of attack. 

\paragraph{Spatial attacks} Spatial attacks are performed by generating rotated and/or translated images with the aim of causing misclassification. Previous work \cite{Engstrom17} has shown that a high number of spatial adversarial examples can be found through a \textit{grid search}. There are infinitely many combinations of rotation and translation possible for a spatial attack, so we attempt to approximate this infinite number by discretizing the continuous space of rotations and translations into a mesh containing all possible combinations of the discretized attack parameters. In particular, we use the following grids in our work. (1) \texttt{grid775}, a grid with 5 values per translation direction (vertical or horizontal) and 31 values for rotation, yielding 775 images; (2) \texttt{grid135}, a grid with 3 values per translation direction and 15 values for rotation, yielding 135 images (for these two grids, we use spatial limits of at most 3px translation in either direction and $30^\circ$ rotation in either direction); (3) \texttt{grid775,10$^\circ$}, the same as \texttt{grid775} but with at most $10^\circ$ of rotation in either direction; (4) \texttt{rot30}, the same as \texttt{grid775} but with no translation; and (5) \texttt{rot10}, the same as \texttt{rot30} but with at most $10^\circ$ of rotation in either direction. 

Attacks (2)--(5) serve as a weaker version of \texttt{grid775} to allow for the appropriate evaluation of the spatial robustness of certain models while avoiding being too strong to break any model claiming to achieve a certain degree of spatial robustness.

\subsection{Datasets}

The base dataset $D$ we use throughout this work is CIFAR-10 \cite{Krizhevsky09}. The datasets $D_R$ and $D_{NR}$ are derived from CIFAR-10 and are freely available as part of \cite{Ilyas2019}. Additionally, we create a custom dataset $D_{mix}$. This dataset is designed such that during training, a certain proportion $\alpha$ of each batch is composed of randomly selected images from $D_R$, while the rest are randomly selected from $D$. One epoch is defined to be over after 50000 images (the size of $D$).

\subsection{Architectures}

\paragraph{Main architectures} In this work we mainly conduct experiments with the ResNet architecture family \cite{He16}, namely ResNet-34 and ResNet-50. The VGG-16 \cite{Simonyan15} model and the ResNet-18 model are used for the purpose of studying the transferability of robustness which is induced by $D_R$ or $D_{mix}$, as the $D_R$ we used are constructed using ResNet-50 in \cite{Ilyas2019}. 

\paragraph{Spatial equivariant networks} To detect the effect of the PGD-robust dataset $D_R$ on special architecture designs aiming to gain equivariance against spatial transformations, we  compare the spatial robust accuracy from spatial equivariant networks trained with the natural dataset $D$ and the robust dataset $D_R$. Our scope of study includes: (a) G-ResNet18 and G-ResNet34 \cite{Cohen16} which use $p4m$ convolutional layers to replace normal 2D convolution layers in the ResNet-18 and ResNet-34 architectures, resulting in feature maps inherently containing rotations in 90$^\circ$ increments. We port the code from \cite{Bielski} in PyTorch into our experiment framework. (b) Spatial Transformer Networks (STNs) \cite{Jaderberg15}, which incorporate a pose predictor module in a backbone ResNet architecture. We adapt the standard STN code from the PyTorch STN tutorial \cite{pytorchstn}. (c) Polar Transformer Networks (PTNs) \cite{Esteves18}, an instantiation of Equivariant Transformer Networks \cite{Tai19} which exploit the rotational equivariance of convolution operations under polar coordinates. We adapt the code offered in \cite{Tai19}.

\subsection{Training details}

\paragraph{Hyperparameters} For experiments about Question \ref{q1} we use batch size 64, a weight decay of $2 \cdot 10^{-4}$, and an initial learning rate of 0.1 which is divided by 10 after 40000 and 60000 iterations for the ResNet-34 model. We train for 100 epochs in total. 

For experiments about Question \ref{q2} and \ref{q3} we use different hyperparameter settings than for Question \ref{q1}. In Question \ref{q2}, the ResNet-50, ResNet-18 and VGG-16 are trained using a batch size of 128, a weight decay of $5 \cdot 10^{-4}$, and an initial learning rate of 0.1 which is divided by 10 every 50 epochs. We train for 150 epochs in total. Due to constraints in computational resources, for Question \ref{q3} we tested two learning rates: 0.1 and 0.01 for the spatial equivariant networks.

\paragraph{Data augmentation} We use two variants of data augmentation. We call the first variant \texttt{std}, which performs a translation of at most 4px in any direction (chosen uniformly at random), performs a random horizontal flip with probability 0.5, jitters the brightness, contrast, and saturation by at most 25\% (chosen uniformly at random), and performs a rotation of at most $2^\circ$ in either direction (chosen uniformly at random). The second variant, \texttt{std*}, is the same as \texttt{std}, but performs a translation of at most $3\px$ in any direction and a rotation of up to $30^\circ$ in either direction.

\subsection{Methods}
For the three questions proposed in Section \ref{sec:intro} we designed a series of experiments to answer them. We describe the detailed methods in the following paragraphs.

\paragraph{Method for Question \ref{q1}} We use a TensorFlow \cite{Abadi2015} framework \cite{Engstrom17,Yang2019} to train a ResNet-34 and evaluate the ASRs for various attacks. In order to examine the claim proposed in \cite{Ilyas2019} that aiming to achieve a high categorical accuracy introduces adversarial vulnerabilities, we design experiments to detect the correlation of model natural accuracies and corresponding ASRs. We train a ResNet-34 model on the CIFAR-10 training set with \texttt{std} data augmentation for 100 epochs. To observe how ASRs for both PGD attacks and spatial attacks evolve with model natural accuracy, we evaluate them every 4 epochs during the training process of the model on the full test set of CIFAR-10. We also evaluate every 625 steps (equals 0.8 epochs) additionally before reaching the fourth epoch, in order to capture the rapid changes in natural accuracy and ASR during the beginning of training. The reason why we use ASR instead of robust accuracy as an evaluation metric is that it reflects how effective an attack method is and it rules out the influence of low natural accuracy of the model during the early training phase by ruling out examples which the model can not even make a correct prediction for in their natural state. 

\paragraph{Method for Question \ref{q2}} We use our framework with \texttt{std} data augmentation to train the ResNet-50, VGG-16, and ResNet-18 on $D_{mix}$ for values of $\alpha \in [0, 1]$ in 0.1-step increments and perform the following PGD attacks: (1) we attack all models with an \ltwo-norm attack constrained by $\epsilon = 0.25$ and 0.5 (with the learning rate and number of iterations described in Section \ref{sec:attacks}); (2) we attack the ResNet-50 with an \linf-norm attack constrained by $\epsilon = \frac{1}{255}, \frac{2}{255}, \frac{4}{255}$, and $\frac{8}{255}$ (with the step size and number of attack steps described in Section \ref{sec:attacks}).

\paragraph{Method for Question \ref{q3}} We use our framework to train the ResNet-50 and the G-ResNet, STN, and PTN (with as backbones ResNet-18 or ResNet-34). Each of the architectures is trained with \texttt{std} as well as \texttt{std*} data augmentation. We perform all the spatial attacks listed in Section \ref{sec:attacks} and log the best robust accuracies for each architecture by searching over the initial learning rates 0.1 and 0.01.

After careful consideration, we discovered that using the algorithm proposed in \cite{Ilyas2019} to construct $D_R$ cannot be used to construct a spatially robust dataset $D_R^S$ due to the fact that doing so would essentially be assigning different labels to the same object in different poses, which would not help with training. We also did not construct a spatially non-robust dataset $D_{NR}^S$ since after a careful grid search over several hyperparameters, we were unable to even achieve the results on $D_{NR}$ listed in \cite{Ilyas2019}.

Our code framework for studying Question \ref{q2} and Question \ref{q3} is built on the PyTorch \cite{Paszke19} \texttt{robustness} library \cite{robustness}.

\section{Results}

\subsection{Answers to Question \ref{q1}}

\begin{figure}[htb]
    \vspace{-7pt}
    \begin{center}
        \includegraphics[width=0.48\textwidth]{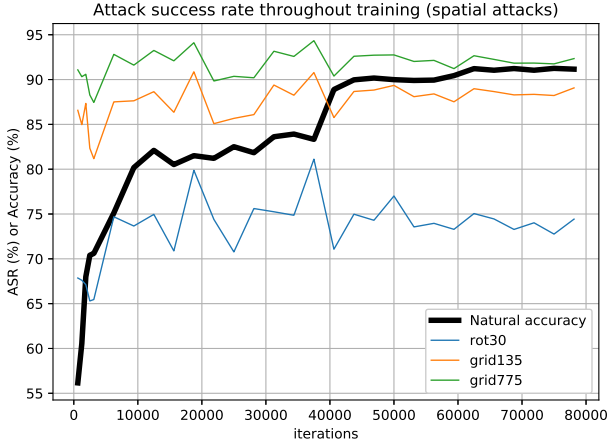}
    \end{center}
    \caption{Attack success rates of spatial attacks at various training stages.} \label{fig:q1_spatial}
    \vspace{-10pt}
\end{figure}

\begin{figure}[htb]
    \vspace{-7pt}
    \begin{center}
        \includegraphics[width=0.48\textwidth]{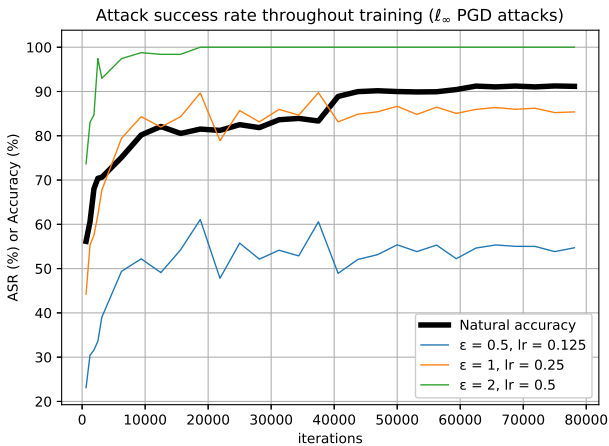}
    \end{center}
    \caption{Attack success rates of \linf PGD attacks at various training stages.} \label{fig:q1_pgd}
    \vspace{-10pt}
\end{figure}

Figures \ref{fig:q1_spatial} and \ref{fig:q1_pgd} show the results of our experiments to determine whether neural networks ``overfit'' to non-robust features during the later stages of training. We distinguish the case of spatial vs. PGD attacks.

\paragraph{For spatial attacks} For spatial attacks (Figure \ref{fig:q1_spatial}), we find that the attack success rate (ASR) fluctuates around a fairly constant rate over the training iterations of the neural network. The ASR for \texttt{rot} remains around 71-80\%, the ASR for \texttt{grid135} remains around 84-90\%, and the ASR for \texttt{grid775} remains around 89-95\%. We conclude that in the case of spatial robustness, neural networks typically do not make a distinction between spatially robust and spatially non-robust features during training, instead using all available features from the very beginning of training.

\paragraph{For PGD attacks} Contrary to the case of spatial attacks, we see a slight upward trend in the ASR when performing $L_{\infty}$-norm PGD attacks as training progresses throughout the early stages of training (natural accuracies 60-80\%), as shown in Figure \ref{fig:q1_pgd}. After reaching a natural accuracy of roughly 80\%, ASR fluctuates around a constant level. We can conclude that at a certain stage of training, the neural network will start using non-robust features to increase natural accuracy.

One possible explanation for the difference in results between spatial and PGD attacks is that spatial attacks are a substantially different class of attacks than PGD attacks. In PGD attacks, there is a constraint based on some norm to limit the attack space, whereas in spatial attacks, the attack space is controlled by a maximum degree of rotation and translation.

In both the spatial and the PGD case, it is clear that neural networks do not specifically over-utilize non-robust features during the later stages of training, as ASR remains fairly constant. We can therefore give the following answer to Question \ref{q1}: \textit{increasing natural accuracy causes higher usage of PGD-non-robust features, shown by ASR increasing at the beginning of training. The current goal of optimizing accuracy to its peak level is therefore partially at fault for introducing adversarial vulnerabilities to PGD attacks caused by non-robust features. In the case of spatial robustness, we cannot draw any meaningful conclusion.}

\begin{table*}[t]
\caption{Accuracies achieved from the experiments for Question \ref{q3}. The first row of the table contains natural accuracies, while the remaining rows contain robust accuracies under the attack mentioned in the leftmost column. Each model has four columns of results: ``$D$'' refers to models trained with the natural dataset, while ``$D_R$'' refers to models trained with the robust dataset. The notation $^*$ indicates the model is trained with \texttt{std*} data augmentation. The entries reported in this table are the best ones across the backbone ResNet-18/-34 and learning rates. The raw experiment data is logged in Tables \ref{tab:stnresults} and \ref{tab:spatialattack_g_resnet}.} \label{tab:spatialattack}
    \begin{adjustwidth}{-.5in}{-.5in}  
        \vspace{15pt}
        \begin{center}
            \begin{tabular}{|c|cccc|cccc|cccc|c|}
                \hline
                & \multicolumn{4}{c|}{ResNet-50} & \multicolumn{4}{c|}{G-ResNet} & \multicolumn{4}{c|}{STN} & PTN\footnotemark \\
                Attack & $D$ & $D_R$ & $D^*$ & $D_R^*$ & $D$ & $D_R$ & $D^*$ & $D_R^*$ & $D$ & $D_R$ & $D^*$ & $D_R^*$ & -- \\ 
                \hline
                None &
                94.84 & 84.75 & 93.9 & 83.53 &  
                \textbf{95.72} & 85.46 & 94.74 & 85.11 &  
                94.05 & 84.39 & 93.12 & 83.72 &   
                -- \\
                \texttt{rot10} &
                80.97 & 67.78 & 83.24 & 69.67 &  
                83.56 & 69.22 & \textbf{87.61} & 73.34 &  
                84.07 & 69.52 & 87.67 & 73.55 &   
                -- \\
                \texttt{rot30} &
                36.77 & 24.94 & 58.75 & 42.62 &  
                41.74 & 26.95 & 75.38 & 51.97 &  
                41.01 & 26.4 & \textbf{86.1} & 70.86 &  
                -- \\
                \texttt{grid775,10$^\circ$} &
                63.34 & 52.3 & 67.78 & 55.31 &  
                68.17 & 55.15 & 75.7 & 60.38 & 
                70.47 & 54.98 & \textbf{77.01} & 60.61 &   
                -- \\
                \texttt{grid135} &
                18.72 & 11.23 & 39.71 & 26.89 &  
                21.84 & 11.81 & 56.06 & 36.64 &  
                22.48 & 11.52 & \textbf{69.47} & 50.76 &   
                -- \\
                \texttt{grid775} &
                15.28 & 9.88 & 35.42 & 24.53 &  
                19.36 & 10.78 & 53.19 & 34.57 &  
                19.94 & 10.09 & \textbf{67.47} & 48.64 &   
                -- \\
                \hline
            \end{tabular}
        \end{center}
    \end{adjustwidth}
    \vspace{-15pt}
\end{table*}

\subsection{Answers to Question \ref{q2}}

\begin{figure}[htb]
    \vspace{-7pt}
    \begin{center}
        \begin{minipage}[b]{0.235\textwidth}
            \includegraphics[width=\textwidth]{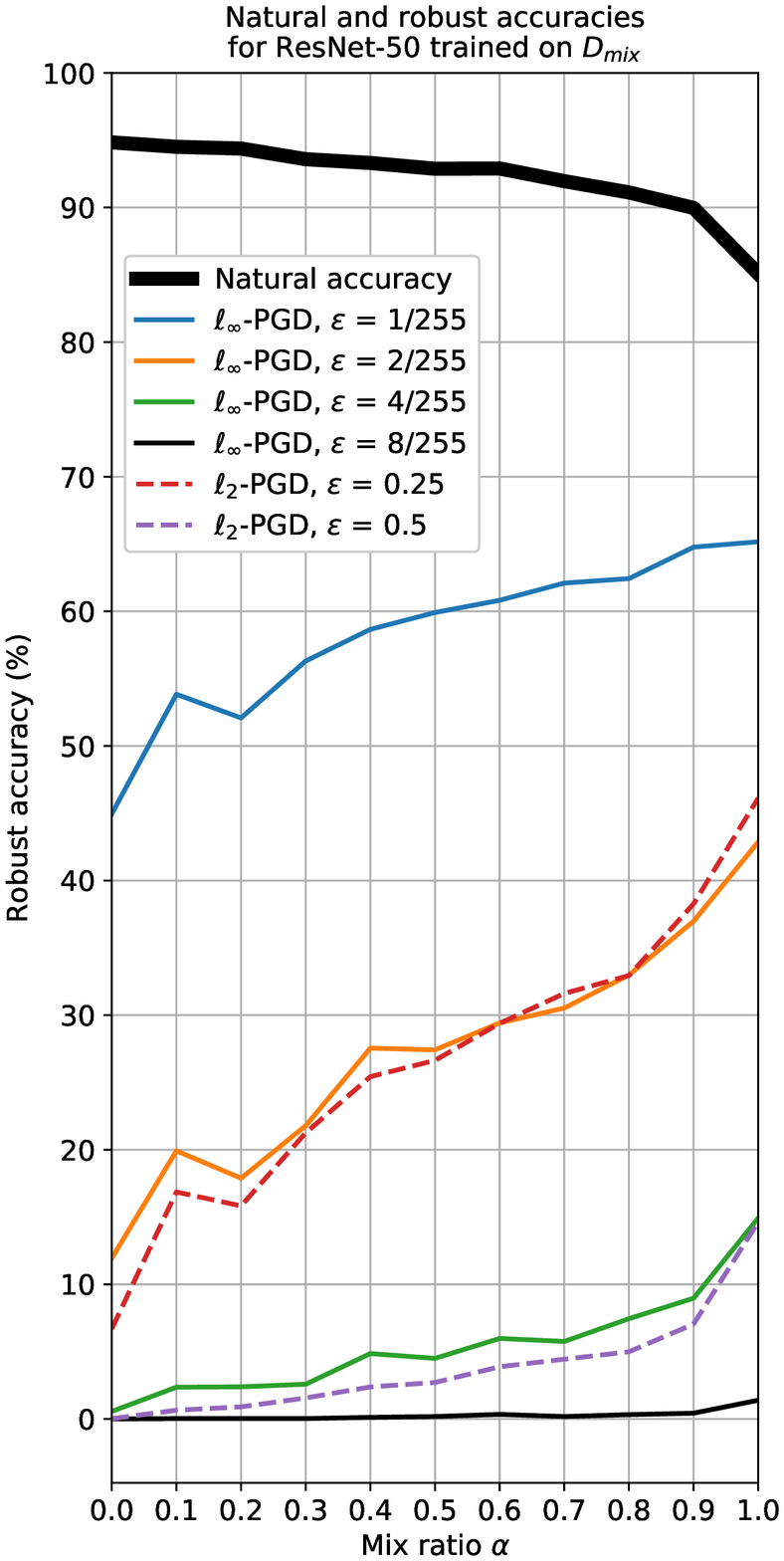}
        \end{minipage}
        \begin{minipage}[b]{0.235\textwidth}
            \includegraphics[width=\textwidth]{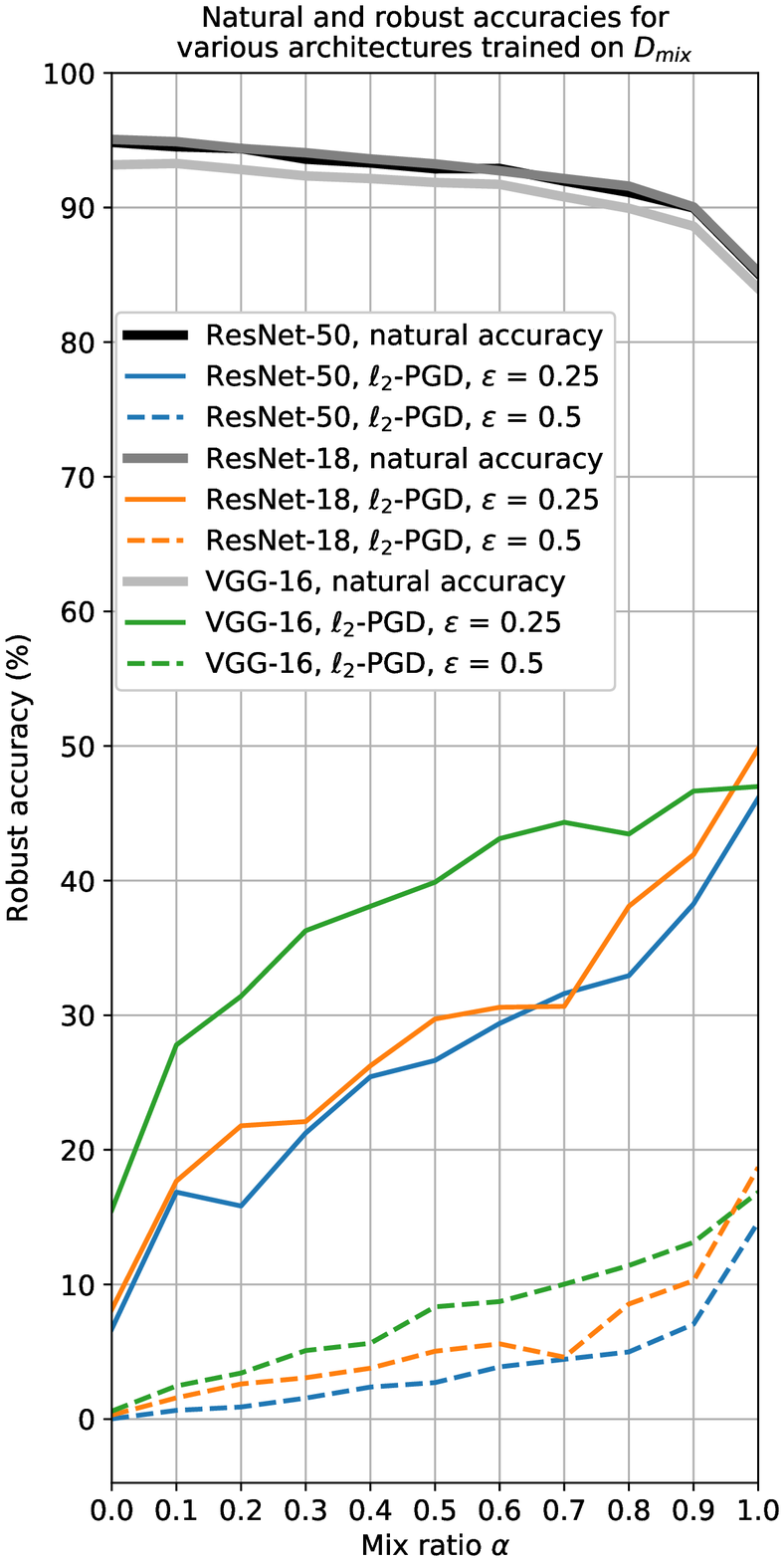}
        \end{minipage}
        \vspace{10pt}
        \caption{Left: Robust accuracies of ResNet-50 trained using $D_{mix}$ for different values of the mix ration $\alpha$, where $\alpha=0$ represents using only natural images and $\alpha=1$ represents using only robust images. Right: Comparison accuracies of the robust accuracies of ResNet-50, ResNet-18, and VGG-16 trained on $D_{mix}$ for different values of $\alpha$. For all attacks, we use the learning rates from Section \ref{sec:attacks}.} \label{fig:mix_results}
    \end{center}
    \vspace{-17pt}
\end{figure}

The results of our experiments on the dataset $D_{mix}$ are summarized in Figure \ref{fig:mix_results}.  

\paragraph{Use of $D_{mix}$ as a defense mechanism} We observe a clear trend that as the mix ration $\alpha$ increases (i.e. the proportion of images from $D_R$ in each training batch increases), the robust accuracy under all considered attacks increases. It is interesting to note that even adding a small amount of images from $D_R$ (e.g. $\alpha=0$ vs. $\alpha=0.1$) greatly increases the robust accuracy, while removing all natural images ($\alpha=0.9$ vs. $\alpha=1$) greatly reduces the natural accuracy. However, even having 10\% of natural images per batch only decreases the natural accuracy by around 4\% compared to using 100\% of natural images. Therefore, using a fixed proportion of images from $D_R$ in each training batch can be considered a fairly effective defense mechanism.

\paragraph{Transferability among different attacks} Ilyas et al. only consider \ltwo-PGD attacks when evaluating their robust dataset $D_R$. Figure \ref{fig:mix_results} (left) shows that $D_R$ is also robust to \linf-PGD attacks (provided the attack is not too strong, as is the case with $\epsilon=\frac{8}{255}$). We conclude that the robustness induced by $D_R$ transfers well among different variations of PGD attack.

\paragraph{Transferability among different architectures} Ilyas et al. generate the robust dataset $D_R$ using a ResNet-50. It is therefore natural to ask whether their generated dataset also has similar robustness-inducing effects on other architectures. Figure \ref{fig:mix_results} (right) shows the results of two different \ltwo-PGD attacks on three different architectures. In all three cases, we observe similar trends to ResNet-50 as $\alpha$ increases. We conclude that a robust dataset generated by one architecture can be used to induce robustness for another architecture.

\subsection{Answers to Question \ref{q3}}
Table \ref{tab:spatialattack} summarizes our results for answering Question \ref{q3}. The models are evaluated under various spatial attack methods. The natural accuracies match baselines in results of \cite{Cohen16, Yang2019}

\paragraph{PGD-robust features do not help boost spatial robustness} Ilyas et al. state that training only with the PGD-robust dataset $D_R$ helps the model to obtain a good degree of robustness in general, but our results clearly demonstrate that this robustness does not apply to the spatial case. In table \ref{tab:spatialattack} we can see that for all architectures, with the same data augmentation procedure, robust accuracy is much lower than its natural counterpart for every spatial attack setting.  

\paragraph{Spatial equivariant networks induce limited spatial robustness with $D_R$} It has been studied in \cite{Yang2019} that several spatial equivariant designs help to boost spatial robustness of the models. However, the results we get from training these specially designed architectures both with $D$ and $D_R$ show that $D_R$ does not further improve the robust accuracy, since all of them do not achieve higher accuracy than when simply trained with $D$.

\footnotetext{For the PTN architecture, due to the training instability for getting the natural accuracy above a meaningful baseline, we omit the results here and leave the results in the appendix.}

\section {Discussion}

Our work reexamined several ideas proposed in \cite{Ilyas2019} under a novel perspective. We try to verify the claimed drawback of the current supervised learning paradigm with a quantitative approach by snapshotting the evolution of the training procedure and measuring ASRs for various attacks. The results we get for Question \ref{q1} show a general trend, but fail to further disentangle the effects of robust and non-robust features on the training process. More experimental designs and a finer-grained control over the training procedure are needed for further study.

Furthermore, we proposed a new approach of constructing a training dataset by mixing natural and robust datasets, which can boost robust accuracy greatly almost without sacrificing natural accuracy. Although it could be considered ``robustness for free'', and it transfers well across attacks under the same class as well as different architectures, the robust accuracy obtained is still below that obtained through state-of-the-art defenses.

Finally, we empirically disproved the possibility that spatial vulnerabilities are caused by PGD-non-robust features, but the reason behind such attacks still remains unknown to us and is out of the scope of this study. 

\section{Conclusion}
In this work, we have studied three questions we raised from the hypothesis of the existence of robust and non-robust features in training data. We first restricted the scope of the study into a smaller set of attacks, by ruling out spatial attacks through monitoring the evolution of ASR with model natural accuracy. We found that the goal of maximizing accuracy is partially responsible for inducing PGD vulnerabilities, but a similar conclusion cannot be drawn for low-dimensional attacks. Furthermore, we found that mixing a certain amount of training examples from the PGD-robust dataset can boost the PGD-robustness by a great degree without substantially hurting natural accuracy, which can be considered as a defense mechanism embedded in the training dataset. This defense transfers well across various first-order attack methods and architectures. Finally, we investigated the relationship between the PGD-robust dataset and spatial equivariant networks and found that it does not help enhance spatial robustness. We speculate that there are no ``spatially robust features'' or ``spatially non-robust features''. This also justifies the no-trade-off phenomenon from \cite{Yang2019}. The spatial attack evaluation results, together with the conclusion for Question \ref{q1}, indicates that we should reexamine certain theories which are presumed to hold for both domains.


\bibliography{references}  
\bibliographystyle{plain}



\newpage

\onecolumn

\section*{\Large{Appendices}}

\begin{appendices}

\section{Omitted experiment results}
\subsection{PTN-ResNet}
\begin{table}[htb]
    \caption{Natural accuracies for training PTN-ResNet18 and PTN-ResNet34 with initial learning rates (lr) = 0.01 or 0.005, trained with different datasets and data augmentations.Each model has four columns of results: ``$D$'' refers to models trained with the natural dataset, while ``$D_R$'' refers to models trained with the robust dataset. The notation $^*$ indicates the model is trained with \texttt{std*} data augmentation. We could not get satisfiable natural accuracy over a reasonable level. Thus, we do not evaluate PTN-ResNet under various spatial attacks and we omit obtained natural accuracy in the main text. We search over initial learning rates {0.01, 0.005} and we log the best results in this table.} \label{tab:ptnresults}
    \begin{adjustwidth}{-.5in}{-.5in}  
        \vspace{15pt}
        \begin{center}
            \begin{tabular}{|c|cccc|cccc|}
                \hline
                  & \multicolumn{4}{c|}{PTN-ResNet18} & \multicolumn{4}{c|}{PTN-ResNet34} \\
         & $D$   & $D_R$        & $D^*$ & $D_R^*$ & $D$          & $D_R$ & $D^*$ & $D_R^*$ \\
         \hline
lr=0.01  & 77.73 & 63.81        & 74.03 & 61.53   & 73.99        & 62.25 & 71.51 & 58.70   \\
lr=0.005 & 77.22 & 66.45        & 75.04 & 62.80   & 74.10        & 61.9  & 70.45 & 58.80  \\
                \hline
            \end{tabular}
        \end{center}
    \end{adjustwidth}
\end{table}

\subsection{STN-ResNet raw experiment results}
\begin{table}[htb]
    \caption{Accuracies achieved with STN-ResNet18 and STN-ResNet34 architectures. The first row of the table contains natural accuracies, while the remaining rows contain robust accuracies under the attack mentioned in the leftmost column. Each model has four columns of results: ``$D$'' refers to models trained with the natural dataset, while ``$D_R$'' refers to models trained with the robust dataset. The notation $^*$ indicates the model is trained with \texttt{std*} data augmentation. We search over initial learning rates {0.01, 0.005} and we log the best results in this table.} \label{tab:stnresults}
    \begin{adjustwidth}{-.5in}{-.5in}  
        \vspace{15pt}
        \begin{center}
            \begin{tabular}{|c|cccc|cccc|}
                \hline
                & 
                \multicolumn{4}{c|}{STN-ResNet-18} & \multicolumn{4}{c|}{STN-ResNet-34} \\
                Attack & $D$ & $D_R$ & $D^*$ & $D_R^*$ & $D$ & $D_R$ & $D^*$ & $D_R^*$ \\ 
                \hline
                None &
                93.55 & 83.75 & 92.83 & 83.05 &  
                94.05 & 84.39 & 93.12 & 83.72   
                \\
                \texttt{rot10} &
                83.54 & 68.66 & 86.46 & 73.08 &  
                84.07 & 69.52 & 87.67 & 73.55   
                \\
                \texttt{rot30} &
                38.66 & 24.98 & 84.72 & 68.93 &  
                41.01 & 26.4 & 86.1 & 70.86   
                \\
                \texttt{grid775,10$^\circ$} &
                68.57 & 54.09 &75.08 &  
                58.47 & 70.47 & 54.98 & 77.01 & 60.61   
                \\
                \texttt{grid135} &
                20.85 & 11.52 & 65.41 & 49.58 &  
                22.48 & 11.04 & 69.47 & 50.76   
                \\
                \texttt{grid775} &
                18.22 & 9.9 & 63.48 & 47.49  
                & 19.94 & 10.09 & 67.47 & 48.64   
                \\
                \hline
            \end{tabular}
        \end{center}
    \end{adjustwidth}
\end{table}

\subsection{G-ResNet raw experiment results}
\begin{table}[htb]
    \caption{Accuracies achieved with G-ResNet18 and G-ResNet34 architectures. The first row of the table contains natural accuracies, while the remaining rows contain robust accuracies under the attack mentioned in the leftmost column. Each model has four columns of results: ``$D$'' refers to models trained with the natural dataset, while ``$D_R$'' refers to models trained with the robust dataset. The notation $^*$ indicates the model is trained with \texttt{std*} data augmentation. We search over initial learning rates {0.1, 0.01}.} \label{tab:spatialattack_g_resnet}
    \begin{adjustwidth}{-.5in}{-.5in}  
        \vspace{15pt}
        \begin{center}
            \begin{tabular}{|c|cccc|cccc|}
                \hline
                & 
                \multicolumn{4}{c|}{G-ResNet-18} & \multicolumn{4}{c|}{G-ResNet-34} \\
                Attack & $D$ & $D_R$ & $D^*$ & $D_R^*$ & $D$ & $D_R$ & $D^*$ & $D_R^*$ \\ 
                \hline
                None &
                94.66 & 85.46 & 94.74 & 85.11 &  
                \textbf{95.72} & 84.93 & 94.72 & 84.55   
                \\
                \texttt{rot10} &
                83.27 & 69.22 & 87.42 & 72.98 &  
                83.56 & 68.8 & \textbf{87.61} & 73.34   
                \\
                \texttt{rot30} &
                41.74 & 26.95 & 69.86 & 51.97 &  
                41.07 & 26.08 & \textbf{75.38} & 46.47   
                \\
                \texttt{grid775,10$^\circ$} &
                68.17 & 54.24 & \textbf{75.7} &  
                58.96 & 67.98 & 55.15 & \textbf{75.7} & 60.38   
                \\
                \texttt{grid135} &
                21.84 & 11.81 & 52.51 & 36.64 &  
                20.99 & 10.99 & \textbf{56.06} & 31.52   
                \\
                \texttt{grid775} &
                19.36 & 10.78 & 49.69 & 34.57  
                & 17.82 & 9.89 & \textbf{53.19} & 29.78   
                \\
                \hline
            \end{tabular}
        \end{center}
    \end{adjustwidth}
\end{table}

\newpage

\begin{figure}[htb]
\section{Training and validation curves while training on $D_{mix}$}
    \vspace{10pt}
    \centering
    \includegraphics[width=0.48\textwidth]{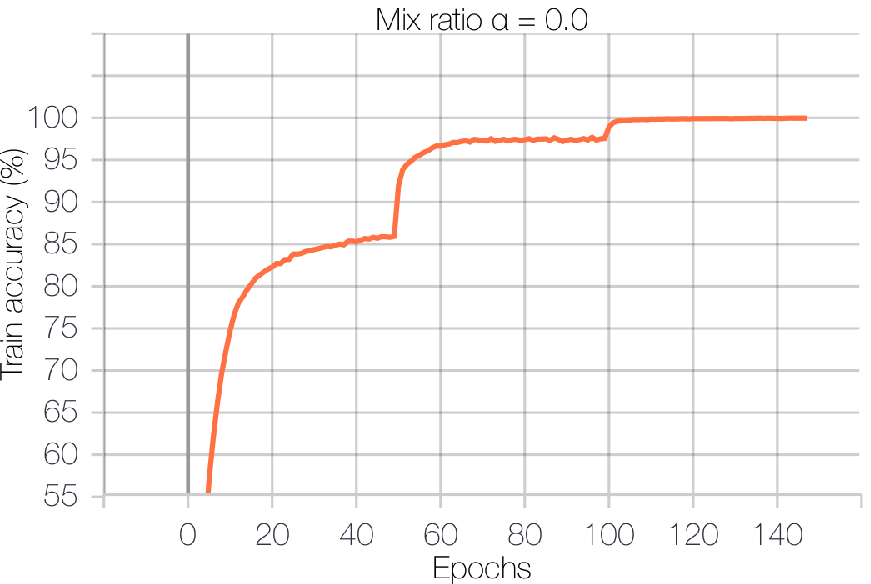}
    \hspace{10pt}
    \includegraphics[width=0.48\textwidth]{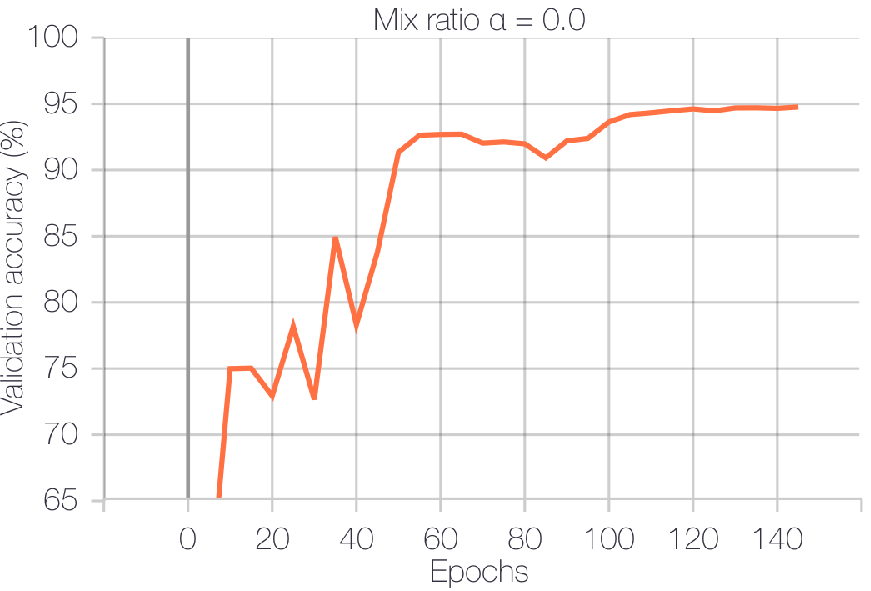}\\
    \vspace{10pt}
    \includegraphics[width=0.48\textwidth]{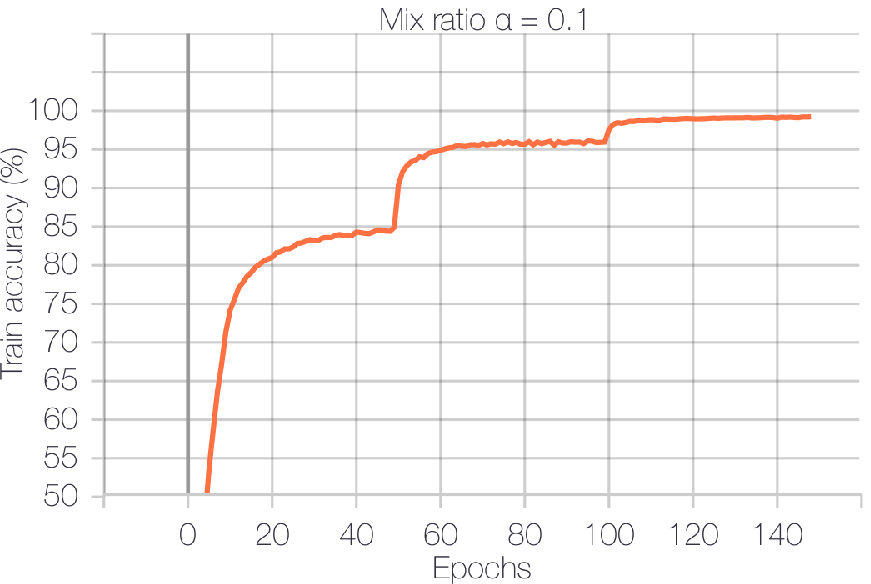}
    \hspace{10pt}
    \includegraphics[width=0.48\textwidth]{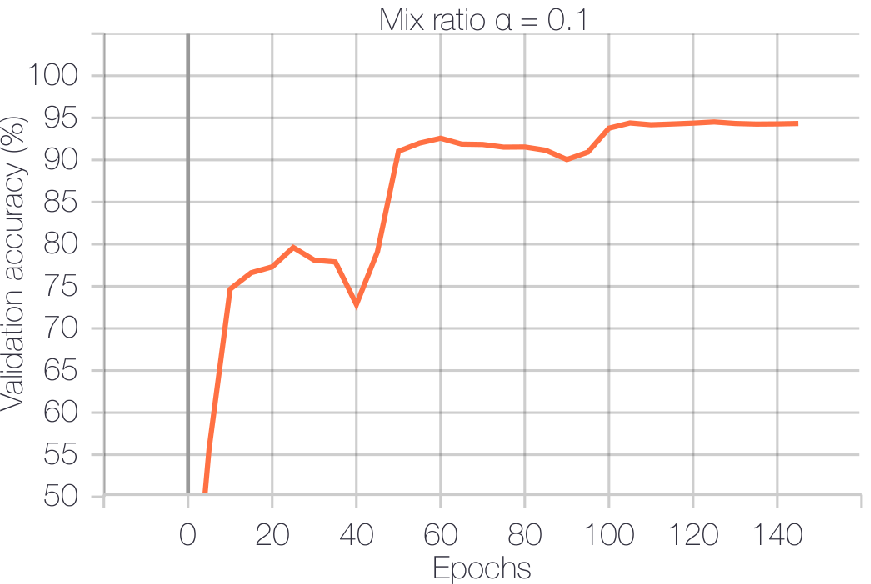}\\
    \vspace{10pt}
    \includegraphics[width=0.48\textwidth]{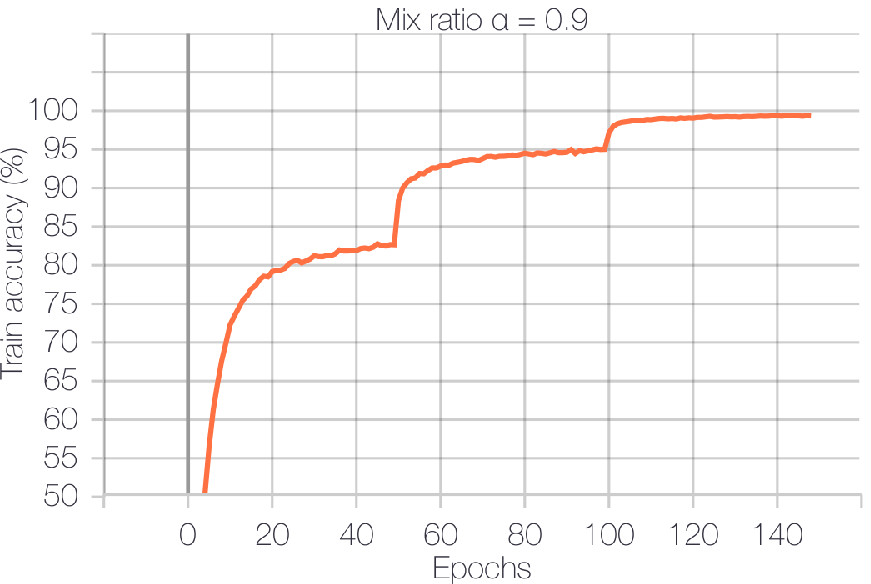}
    \hspace{10pt}
    \includegraphics[width=0.48\textwidth]{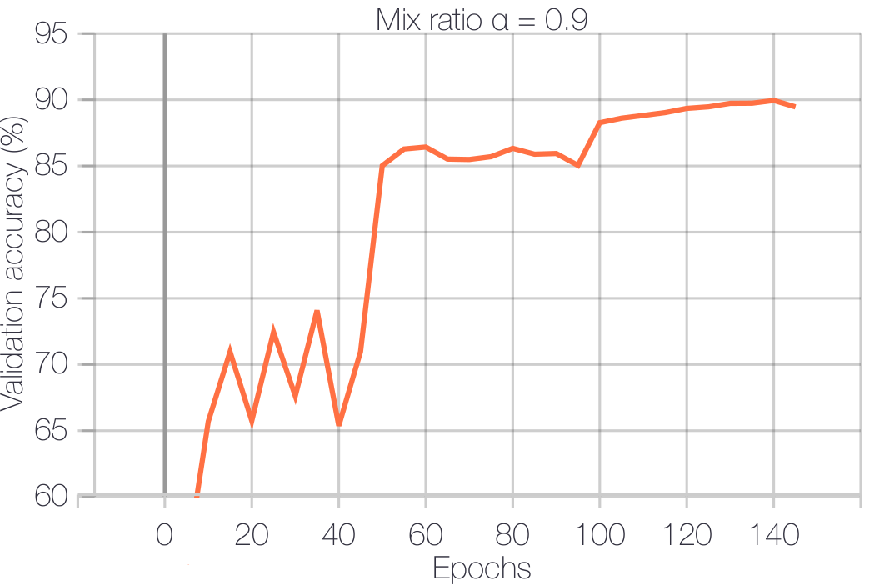}\\
    \vspace{10pt}
    \includegraphics[width=0.48\textwidth]{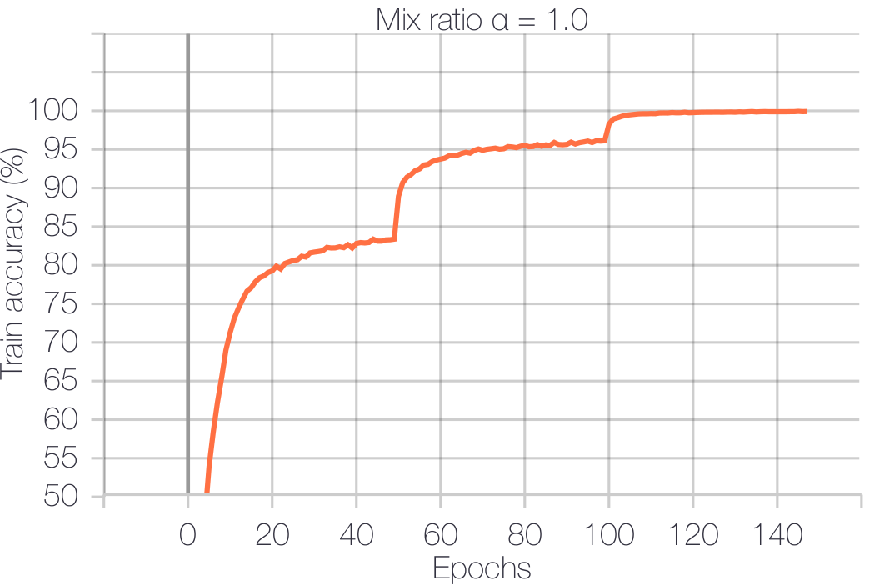}
    \hspace{10pt}
    \includegraphics[width=0.48\textwidth]{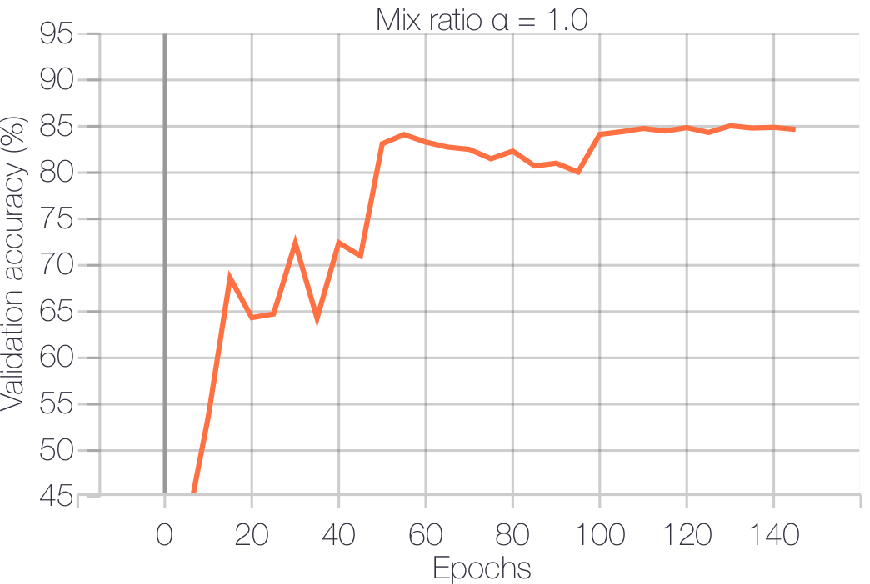}\\
\end{figure}

\end{appendices}
\end{document}